\documentclass{article}

\PassOptionsToPackage{numbers, compress}{natbib}

\usepackage[nonatbib, preprint]{neurips_2023}

\usepackage[utf8]{inputenc} 
\usepackage[T1]{fontenc}    
\usepackage{hyperref}       
\usepackage{url}            
\usepackage{booktabs}       
\usepackage{amsfonts}       
\usepackage{nicefrac}       
\usepackage{microtype}      
\usepackage{xcolor}         
\usepackage{tcolorbox}
\usepackage{cite}
\usepackage{color}

\usepackage{marvosym}
\newcommand{\authorskip}{\hspace{2.5mm}}

\title{MultiModal-GPT: A Vision and Language Model for Dialogue with Humans}

\author{Tao Gong$^{1*}$ \authorskip Chengqi Lyu$^{1*}$ \authorskip Shilong Zhang$^{2,1*}$ \authorskip Yudong Wang$^{1,3*}$
\authorskip Miao Zheng$^{1*}$ \And Qian Zhao$^{1*}$ \authorskip Kuikun Liu$^{1*}$
\authorskip Wenwei Zhang$^{1*}$ \authorskip Ping Luo$^{2,1}$ \authorskip Kai Chen$^{1\textrm{\Letter}}$ \\[2mm]
\small $^{*}$equal contribution, in random order \\[2mm]
$^{1}$Shanghai AI Laboratory \authorskip $^{2}$The University of Hong Kong \\
$^{3}$School of Electrical and Information Engineering, Tianjin University \\
{\tt\small $\left \{\texttt{gongtao, lvchengqi, zhangshilong, wangyudong, zhengmiao}\right\}$@pjlab.org.cn} \\
{\tt\small $\left \{\texttt{zhaoqian, liukuikun, zhangwenwei, chenkai}\right\}$@pjlab.org.cn}
}

\begin{document}

\maketitle

\begin{abstract}
  We present a vision and language model named MultiModal-GPT to conduct multi-round dialogue with humans. MultiModal-GPT is capable of following diverse instructions, such as generating detailed captions, counting specific objects, and addressing general inquiries posed by users. The model is efficiently fine-tuned from OpenFlamingo, with Low-rank Adapter (LoRA) incorporated in both the gated-cross-attention and self-attention components of the language model.
  Our approach involves constructing instruction templates that incorporate vision and language data for multi-modality instruction tuning, enabling the model to comprehend and adhere to human directives. We observe that the quality of training data is crucial for effective dialogue performance, as a limited dataset with short responses may cause the model to generate brief replies to any instruction.
  To further enhance MultiModal-GPT's conversational abilities, we employ language-only instruction-following data for joint training alongside visual-language instructions. Utilizing the \emph{same} instruction template for both types of data results in a significant improvement in dialogue performance. Our experiments demonstrate MultiModal-GPT's proficiency in maintaining continuous dialogues with humans. The code, dataset, and demo can be found at \href{https://github.com/open-mmlab/Multimodal-GPT}{https://github.com/open-mmlab/Multimodal-GPT}.
\end{abstract}

\section{Introduction}

Humans interact with the world through multiple channels, including vision and language, each of which has a unique advantage in representing and conveying certain concepts of the world, thus contributing to a better understanding of the world. A central objective of artificial intelligence research is to create a versatile assistant capable of effectively following multimodal vision-and-language instructions that align with human intentions, in order to accomplish a diverse array of real-world tasks.

Recently, GPT-4 \cite{gpt4} has demonstrated remarkable proficiency in multi-modal dialogues with humans. Although GPT-4's \cite{gpt4} exceptional capabilities have been observed, the mechanisms underpinning its outstanding performance remain elusive. Studies such as Mini-GPT4 \cite{zhu2023minigpt} and LLaVA \cite{liu2023visual} have sought to replicate this performance by aligning visual representations with the input space of LLM, subsequently utilizing the original self-attention in the LLM to process visual information. However, incorporating such models with detailed or spatiotemporal visual information can be computationally intensive due to the potentially large number of image tokens. Furthermore, both models employ vicuna~\cite{vicuna2023}, an open-source chatbot refined through fine-tuning LLaMA~\cite{touvron2023llama} on user-generated conversations from ChatGPT, which omits the language instruction tuning phase in their research.


To address these challenges, we build upon the open-source Flamingo framework \cite{alayrac2022flamingo}, a multimodal pre-trained model that deploys a perceiver resampler to efficiently extract visual information from the vision encoder, while also employing gated cross-attention layers for image-text interactions. This model has been pre-trained on an extensive dataset of image-text pairs, showcasing robust few-shot visual comprehension capabilities. Nevertheless, it lacks the capacity to engage in zero-shot multi-turn image-text dialogues. As a result, our goal is to fine-tune OpenFlamingo using comprehensive datasets of image and text instructions, enabling the model to conduct conversations that more closely align with human preferences. By capitalizing on OpenFlamingo's foundational strengths, we aspire to narrow the performance gap between the model's existing capabilities and the desired outcome of more accurate, human-like interactions in multimodal dialogues. We have dubbed our multimodal chatbot MultiModal-GPT.

We also use a unified instruction template for both language and visual instruction data during model training. We first construct instruction templates with vision and language data to train the MultiModal-GPT. We find the training data is vital with respect to the performance of the MultiModal-GPT. Some datasets, such as VQA v2.0 \cite{goyal2017making}, OKVQA \cite{marino2019ok}, GQA \cite{hudson2019gqa}, CLEVR \cite{johnson2017clevr} and NLVR \cite{suhr2017corpus} datasets, will degrade the dialogue performance of the MultiModal-GPT, since the response in these datasets is restricted to one or two words (e.g., yes/no). Consequently, when these datasets are incorporated into the training process, the model exhibits a tendency to generate answers comprising merely one or two words. This brevity is not conducive to user-friendliness.

To further enhance the ability of MultiModal-GPT to chat with people, we also collect language data and define a unified instruction template to jointly train the MultiModal-GPT. The joint training of language-only instructions and visual and language instructions effectively improves the performance of the model. We show various demos to show the ability of continuous dialogue of MultiModal-GPT with humans.

\section{Unified Instruction Template}

We propose a unified template for the integration of unimodal linguistic data and multimodal vision-and-language data, with the objective of effectively training the MultiModal-GPT model in a synergistic manner. This unified approach aims to enhance the model's performance across diverse tasks by leveraging the complementary strengths of both data modalities and fostering a more profound understanding of the underlying concepts.

\subsection{Language-only Instruction Template}

\begin{table}[!h]
\begin{tcolorbox}
\textcolor[rgb]{0,0.7,0}{<BOS>} Below is an instruction that describes a task. Write a response that appropriately completes the request 

\#\#\# Instruction: \textcolor[rgb]{0,0,0.8}{\{instruction\}}

\#\#\# Input: \textcolor[rgb]{0,0,0.8}{\{input\}}

\#\#\# Response: \textcolor[rgb]{0.8,0,0}{\{response\}}  \textcolor[rgb]{0.8,0,0}{<EOS>}

\end{tcolorbox}
\caption{The input sequence of language data used to train the model. The \textcolor[rgb]{0,0,0.8}{\{instruction\}}, \textcolor[rgb]{0,0,0.8}{\{input\}} and \textcolor[rgb]{0.8,0,0}{\{response\}} are texts from the source data. Only the \textcolor[rgb]{0.8,0,0}{\{response\}} part and \textcolor[rgb]{0.8,0,0}{<EOS>} token will be calculated loss.}
\label{lan_prompt}
\end{table}

We employ the Dolly 15k and Alpaca GPT4 datasets \cite{peng2023instruction} as resources for assessing language-only instruction-following capabilities. These datasets have been specifically designed to improve the performance of language models in executing instruction-based tasks. To ensure consistent instruction-following format, we utilize the prompt template presented in Table~\ref{lan_prompt} for structuring the dataset input.

\subsection{Vision and Language Instruction Template}

\begin{table}[!h]
\begin{tcolorbox}
\textcolor[rgb]{0,0.7,0}{<BOS>} Below is an instruction that describes a task. Write a response that appropriately completes the request 

\#\#\# Image: \textcolor[rgb]{0,0,0.8}{<image\_token>}\ 

\#\#\# Instruction: \textcolor[rgb]{0,0,0.8}{\{question\}}

\#\#\# Response: \textcolor[rgb]{0.8,0,0}{\{response\}}\textcolor[rgb]{0.8,0,0}{<EOS>}

\#\#\# Instruction: \textcolor[rgb]{0,0,0.8}{\{question\}}

\#\#\# Response: \textcolor[rgb]{0.8,0,0}{\{response\}} \textcolor[rgb]{0.8,0,0}{<EOS>}\ 
\end{tcolorbox}
\caption{The input sequence of vision and language data used to train the model. The \textcolor[rgb]{0,0,0.8}{\{question\}} and \textcolor[rgb]{0.8,0,0}{\{response\}} are texts from the source data.  \textcolor[rgb]{0,0,0.8}{<image\_token>} is a token denoting the existence of image. Note that there are multi-round dialogues if the dataset has. Only the \textcolor[rgb]{0.8,0,0}{\{response\}} part and \textcolor[rgb]{0.8,0,0}{<EOS>} token will be calculated loss.}
\label{vision_prompt}
\end{table}

We utilize a diverse selection of vision and language instruction-following datasets in our study, including LLaVA \cite{liu2023visual}, Mini-GPT4 \cite{zhu2023minigpt}, A-OKVQA \cite{schwenk2022okvqa}, COCO Caption \cite{karpathy2015deep}, and OCR VQA \cite{mishraICDAR19}. These datasets encompass a wide array of applications and domains, thereby facilitating the comprehensive evolving of our model's performance.

In order to present the text in a consistent, instruction-following format, we adopt the prompt delineated in Table~\ref{vision_prompt} as a template for structuring these datasets. By adhering to a standardized format, we ensure that our model is better equipped to process the information and respond accordingly.

It is important to note that the COCO Caption dataset generally does not include instructional content, as it predominantly consists of descriptive captions. To overcome this limitation and incorporate instructional data, we employ the GPT-4 \cite{gpt4} model to generate pertinent instructions for the COCO Caption dataset. This integration of synthesized instructions enriches the dataset, enabling our model to achieve a more robust capabilities in processing and responding to human instructions.

Table~\ref{tab:caption} showcases a variety of examples illustrating the instructions generated for the COCO Caption dataset, demonstrating the effectiveness of our approach in adapting the dataset to better suit our research objectives.

\begin{table}
\begin{tcolorbox}
\begin{itemize}
\item Can you describe the image?
\item Could you provide a description of the image?
\item What do you see in this image?
\item Share your thoughts on the content of the image.
\item Please narrate what's happening in the picture.
\item Can you give a brief explanation of the image?
\item Describe the main elements and details present in the image.
\item In your own words, what is depicted in the image?
\item How would you describe the image's content in a caption?
\item Can you suggest an insightful caption that highlights the underlying message of the image?
\end{itemize}
\end{tcolorbox}
\caption{The list of instructions for image caption.}
\label{tab:caption}
\end{table}

\section{Method}

\subsection{Architecture}

\begin{figure}[!t]
  \centering
  \includegraphics[width=\textwidth]{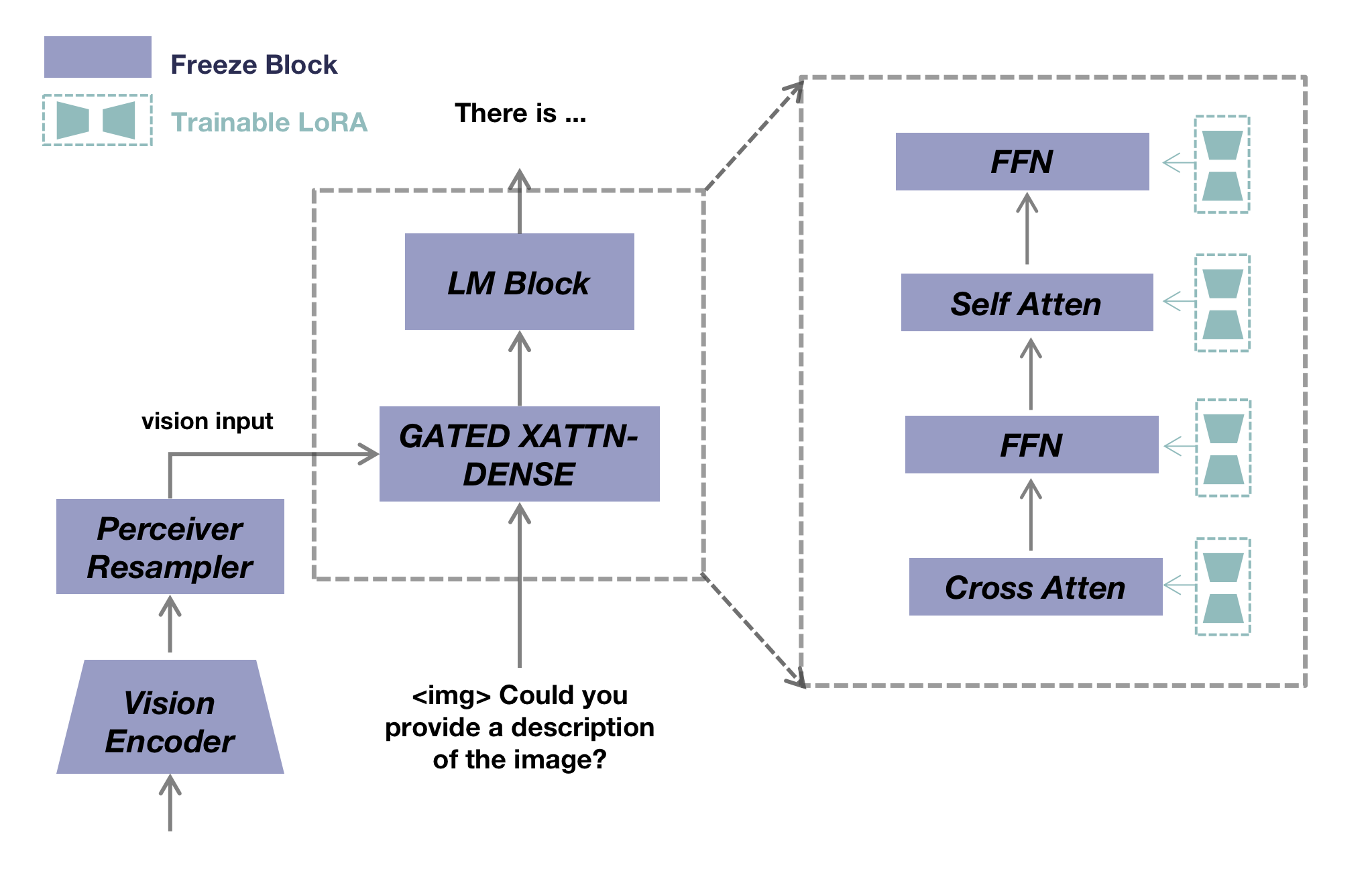}
  \caption{The overall framework of MultiModal-GPT. MultiModal-GPT consists of a vision encoder, a perceiver resampler to receive the spatial features from the vision encoder, and a language decoder which is conditioned on the spatial features from the perceiver resampler by cross-attention in order to encode the feature of vision into text. We freeze the whole open-flamingo model and add LoRA to the self-attention part, the cross-attention part, and the FFN part in the language decoder to finetune MultiModal-GPT.}
  \label{model_architechture}
  
\end{figure}

The proposed MultiModal-GPT is based on the open-flamingo model \cite{alayrac2022flamingo}. As shown in Figure \ref{model_architechture}, MultiModal-GPT consists of a vision encoder from CLIP \cite{radford2021learning}, a perceiver resampler to receive the spatial features from the vision encoder, and a language decoder LLaMA \cite{touvron2023llama}. Note that the language decoder is conditioned on the spatial features from the perceiver resampler by cross-attention in order to encode the feature of vision into text. Please refer to \cite{alayrac2022flamingo} for more details of the model architecture.

\subsection{Joint Training}
We use both language-only instruction-following data and vision and language instruction-following data to train the MultiModal-GPT jointly. As shown in Fig.\ref{model_architechture}, We freeze the whole open-flamingo model and add LoRA \cite{hu2021lora} to the self-attention, cross-attention, and FFW part in the language decoder to finetune MultiModal-GPT. The MultiModal-GPT is trained by predicting the next token of the text, and only the \textcolor[rgb]{0.8,0,0}{\{response\}} and \textcolor[rgb]{0.8,0,0}{<EOS>} tokens in the input sequence are involved in the loss calculation. 

\section{Experiments}

\subsection{Implementation Details}

We jointly train the MultiModal-GPT model using a comprehensive mix of language data and vision and language data sources to enhance its performance. The language datasets include Dolly 15k and Alpaca GPT4 \cite{peng2023instruction}, while the vision and language datasets encompass LLaVA \cite{liu2023visual}, Mini-GPT4 \cite{zhu2023minigpt}, A-OKVQA \cite{schwenk2022okvqa}, COCO Caption \cite{karpathy2015deep}, and OCR VQA \cite{mishraICDAR19}. Other vision and language instruction datasets, such as MultiInstruct \cite{xu2022multiinstruct} can also be explored, and we left it as future work. This combination of datasets aims to provide a diverse and rich training environment for the MultiModal-GPT model.

To effectively train the model, we incorporate the entire text corpus from the Dolly 15k and Alpaca GPT4 datasets. Similarly, we include all image-text pairs available from the LLaVA and Mini-GPT4 datasets to ensure adequate exposure to various contexts and situations. However, the quality of the A-OKVQA, COCO Caption, and OCR VQA datasets is considered inferior compared to LLaVA and Mini-GPT4. To account for this disparity while still benefiting from the additional data, we include a random sample of 5000 image-text pairs from the A-OKVQA dataset and 512 image-text pairs each from the COCO Caption and OCR VQA datasets in the training process.

To train the model, we utilize 8 A100 GPUs and complete the training process within a single epoch. The batch size per GPU for both vision and language instruction following data and language-only instruction following data is set to 1. We use gradient accumulation, and update the parameters of the LoRA every 16 iterations. Each iteration encompasses one vision-language pair and one language-only instruction data. Consequently, the aggregate batch size amounts to 256. We employ a learning rate of 1e-5, using a cosine learning rate scheduler to adjust the learning rate during the training process.

\subsection{The Quality of Data Matters}

In order to enhance the diversity of vision and language data, we initially incorporated a multitude of datasets in our preliminary experiments. However, we observed that the quality of certain datasets, including VQA v2.0 \cite{goyal2017making}, OKVQA \cite{marino2019ok}, GQA \cite{hudson2019gqa}, CLEVR \cite{johnson2017clevr}, and NLVR \cite{suhr2017corpus}, was suboptimal. This is primarily due to the fact that the \textcolor[rgb]{0,0,0.8}{\{response\}} in these datasets is restricted to one or two words (e.g., yes/no). Consequently, when these datasets are incorporated into the training process, the model exhibits a tendency to generate answers comprising merely one or two words. This brevity is not conducive to user-friendliness. As a result, we have opted to exclude these datasets from the final version of our study.

\subsection{Demos}
\begin{figure}
  \centering
  \includegraphics[width=1\textwidth]{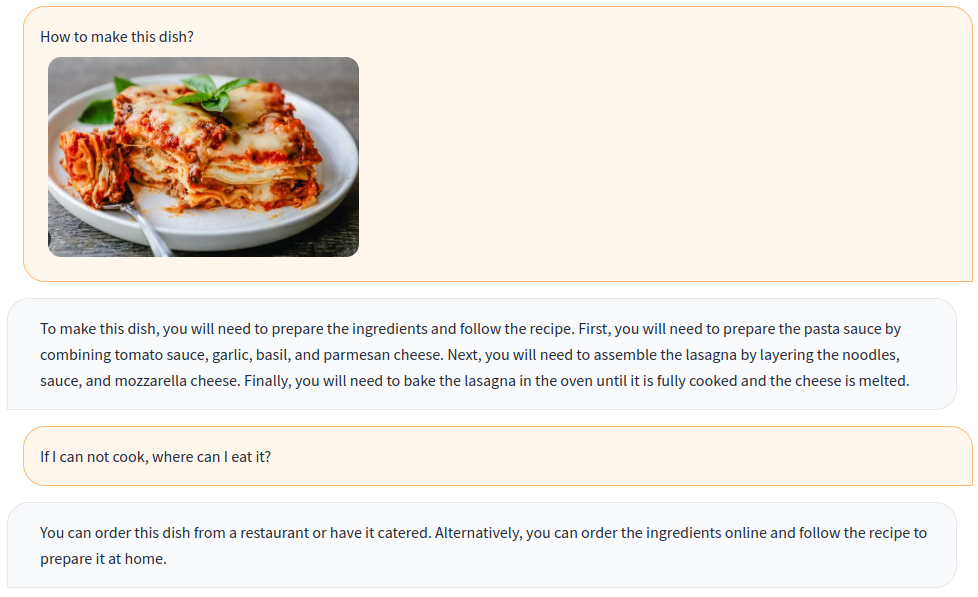}
  \caption{The MultiModal-GPT can give a recipe to bake lasagna, and tell users where to eat it.}
  \label{lasagna}
\end{figure}

We present a variety of demos that exhibit the capabilities of MultiModal-GPT in engaging in conversation with humans. As illustrated in Figure \ref{lasagna}, MultiModal-GPT can successfully identify lasagna and provide a comprehensive recipe for preparing the dish when prompted by the user in the initial dialogue. In the subsequent dialogue, MultiModal-GPT offers reasonable suggestions on where to dine when the user inquires about potential eateries. This exemplifies the model's capacity for maintaining an ongoing dialogue.

\begin{figure}
  \centering
  \includegraphics[width=1\textwidth]{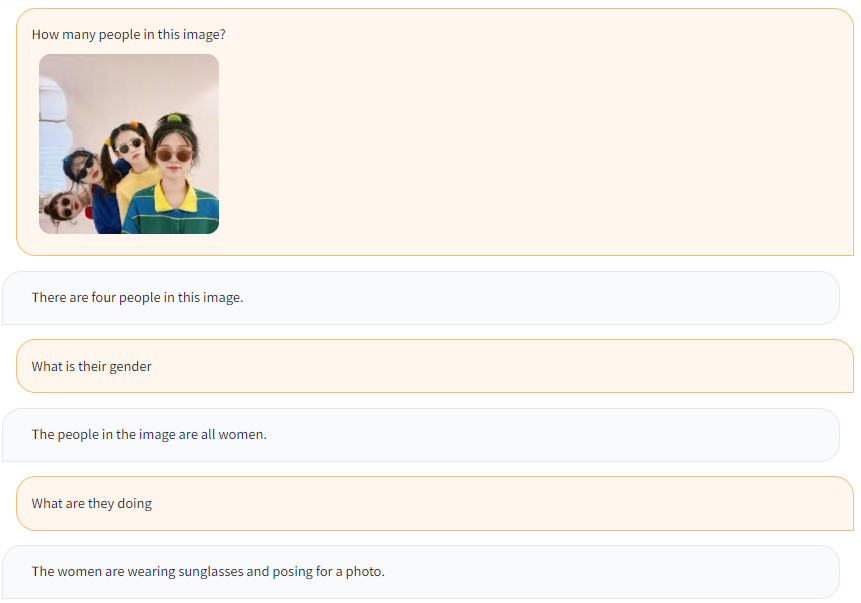}
  \caption{The MultiModal-GPT knows there are 4 women in the image, and knows what they are doing.}
  \label{4women}
\end{figure}

\begin{figure}
  \centering
  \includegraphics[width=0.8\textwidth]{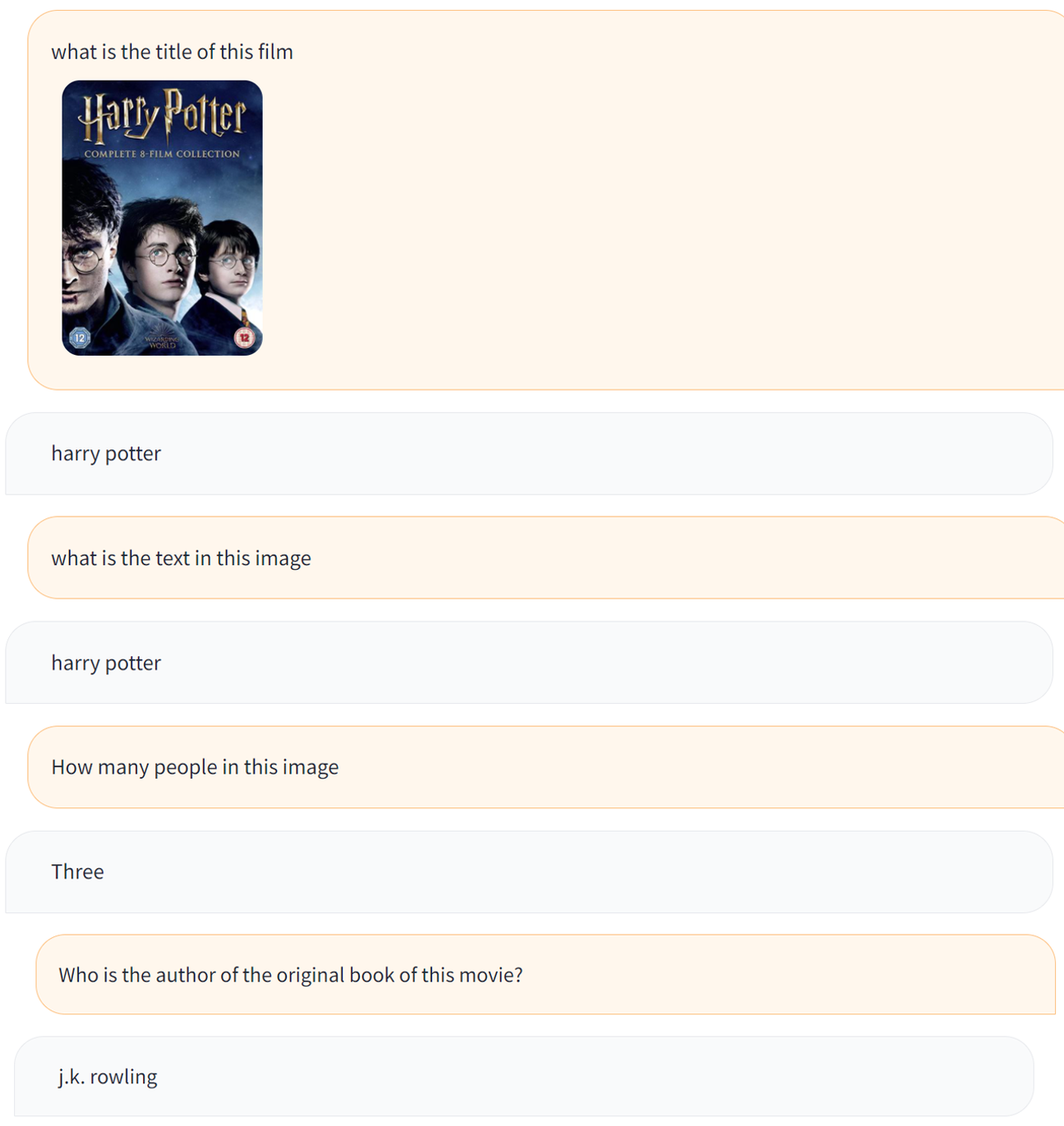}
  \caption{The MultiModal-GPT can recognize the characters in the image and knows the author of the book.}
  \label{harrypotter_ocr}
\end{figure}

\begin{figure}
  \centering
  \includegraphics[width=1\textwidth]{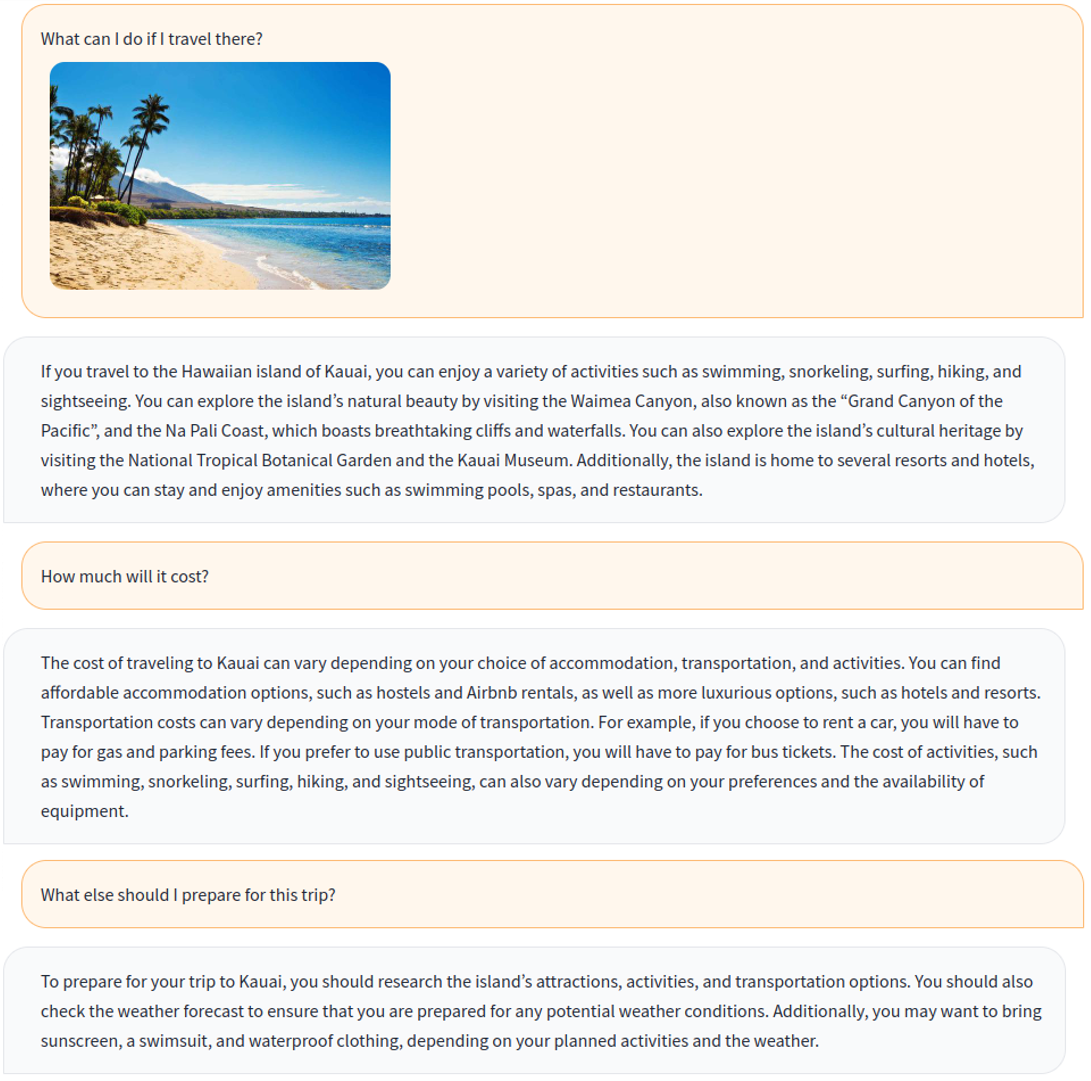}
  \caption{The MultiModal-GPT can answer general questions about the traveling of users.}
  \label{travel}
\end{figure}

Additionally, Figure \ref{4women} demonstrates MultiModal-GPT's counting abilities, while Figure \ref{harrypotter_ocr} showcases its OCR capabilities. MultiModal-GPT is also adept at addressing general inquiries regarding users' travel plans, as depicted in Figure \ref{travel}. More demos can be found at \href{https://github.com/open-mmlab/Multimodal-GPT}{https://github.com/open-mmlab/Multimodal-GPT}.

\begin{figure}
  \centering
  \includegraphics[width=1\textwidth]{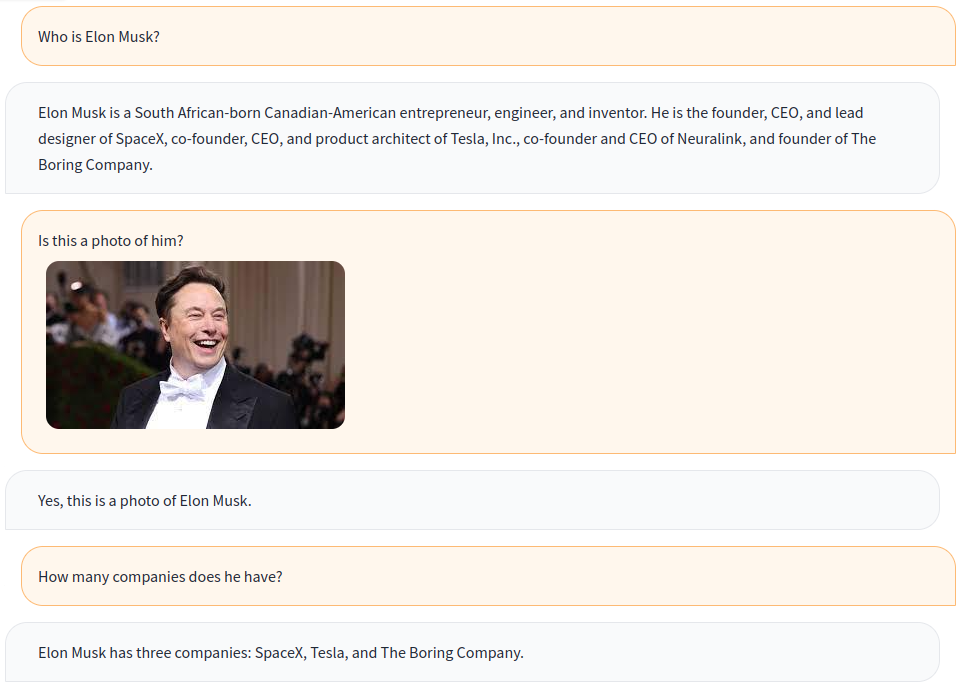}
  \caption{The MultiModal-GPT can recognize the image of Elon Musk, and answer questions about Elon Musk.}
\end{figure}

\begin{figure}
  \centering
  \includegraphics[width=1\textwidth]{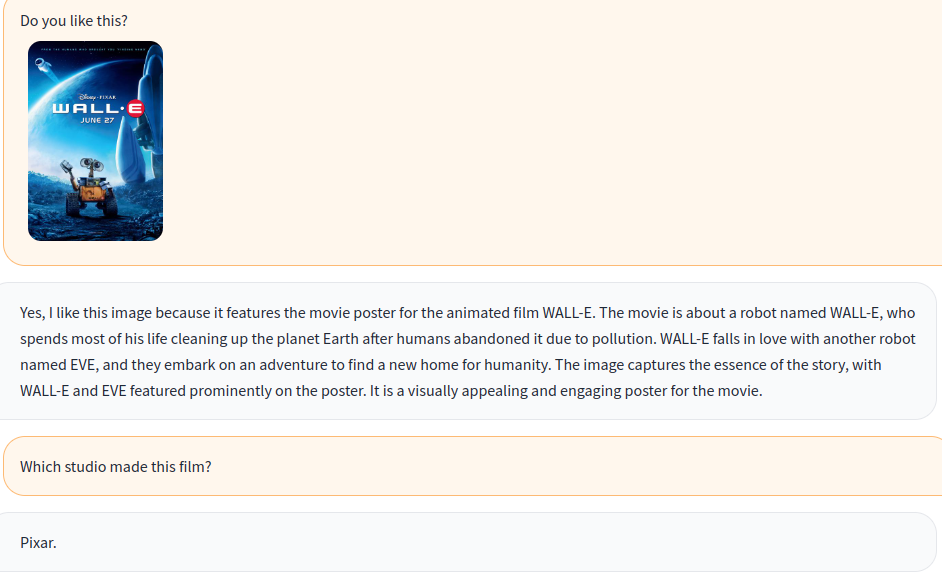}
  \caption{The MultiModal-GPT knows the film and knows which studio made the film.}
\end{figure}

\begin{figure}
  \centering
  \includegraphics[width=0.9\textwidth]{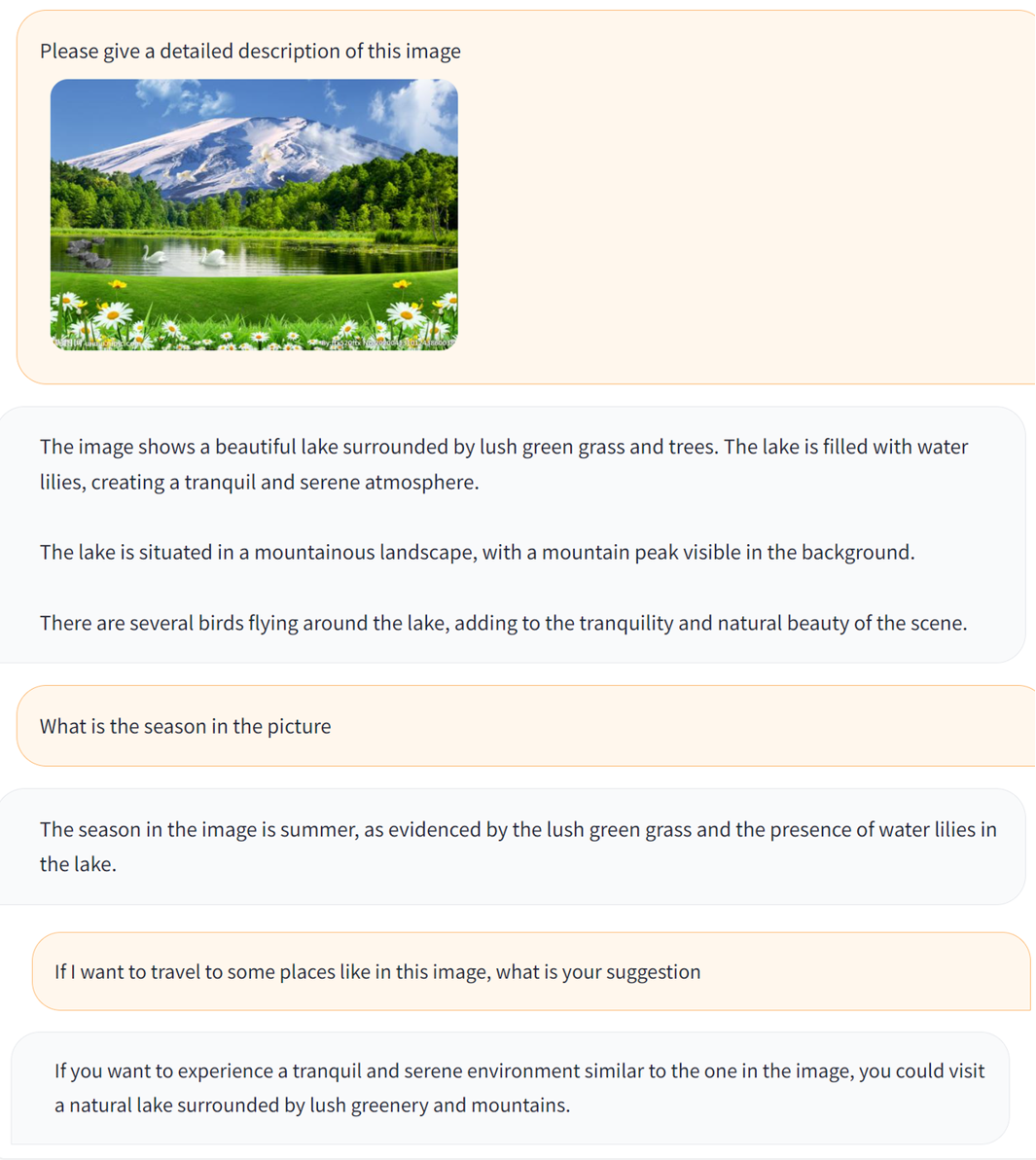}
  \caption{The MultiModal-GPT can generate a detailed description for the image and has the ability to reason the season in the image.}
  \label{caption}
\end{figure}

\clearpage

\bibliographystyle{ieee_fullname}
\bibliography{egbib}

\begin{thebibliography}{10}\itemsep=-1pt

\bibitem{alayrac2022flamingo}
Jean-Baptiste Alayrac, Jeff Donahue, Pauline Luc, Antoine Miech, Iain Barr,
  Yana Hasson, Karel Lenc, Arthur Mensch, Katherine Millican, Malcolm Reynolds,
  et~al.
\newblock Flamingo: a visual language model for few-shot learning.
\newblock {\em Advances in Neural Information Processing Systems},
  35:23716--23736, 2022.

\bibitem{vicuna2023}
Wei-Lin Chiang, Zhuohan Li, Zi Lin, Ying Sheng, Zhanghao Wu, Hao Zhang, Lianmin
  Zheng, Siyuan Zhuang, Yonghao Zhuang, Joseph~E. Gonzalez, Ion Stoica, and
  Eric~P. Xing.
\newblock Vicuna: An open-source chatbot impressing gpt-4 with 90\%* chatgpt
  quality, March 2023.

\bibitem{goyal2017making}
Yash Goyal, Tejas Khot, Douglas Summers-Stay, Dhruv Batra, and Devi Parikh.
\newblock Making the v in vqa matter: Elevating the role of image understanding
  in visual question answering.
\newblock In {\em Proceedings of the IEEE conference on computer vision and
  pattern recognition}, pages 6904--6913, 2017.

\bibitem{hu2021lora}
Edward~J Hu, Yelong Shen, Phillip Wallis, Zeyuan Allen-Zhu, Yuanzhi Li, Shean
  Wang, Lu Wang, and Weizhu Chen.
\newblock Lora: Low-rank adaptation of large language models.
\newblock {\em arXiv preprint arXiv:2106.09685}, 2021.

\bibitem{hudson2019gqa}
Drew~A Hudson and Christopher~D Manning.
\newblock Gqa: A new dataset for real-world visual reasoning and compositional
  question answering.
\newblock In {\em Proceedings of the IEEE/CVF conference on computer vision and
  pattern recognition}, pages 6700--6709, 2019.

\bibitem{johnson2017clevr}
Justin Johnson, Bharath Hariharan, Laurens Van Der~Maaten, Li Fei-Fei, C
  Lawrence~Zitnick, and Ross Girshick.
\newblock Clevr: A diagnostic dataset for compositional language and elementary
  visual reasoning.
\newblock In {\em Proceedings of the IEEE conference on computer vision and
  pattern recognition}, pages 2901--2910, 2017.

\bibitem{karpathy2015deep}
Andrej Karpathy and Li Fei-Fei.
\newblock Deep visual-semantic alignments for generating image descriptions.
\newblock In {\em Proceedings of the IEEE conference on computer vision and
  pattern recognition}, pages 3128--3137, 2015.

\bibitem{liu2023visual}
Haotian Liu, Chunyuan Li, Qingyang Wu, and Yong~Jae Lee.
\newblock Visual instruction tuning.
\newblock {\em arXiv preprint arXiv:2304.08485}, 2023.

\bibitem{marino2019ok}
Kenneth Marino, Mohammad Rastegari, Ali Farhadi, and Roozbeh Mottaghi.
\newblock Ok-vqa: A visual question answering benchmark requiring external
  knowledge.
\newblock In {\em Proceedings of the IEEE/cvf conference on computer vision and
  pattern recognition}, pages 3195--3204, 2019.

\bibitem{mishraICDAR19}
Anand Mishra, Shashank Shekhar, Ajeet~Kumar Singh, and Anirban Chakraborty.
\newblock Ocr-vqa: Visual question answering by reading text in images.
\newblock In {\em ICDAR}, 2019.

\bibitem{gpt4}
OpenAI.
\newblock Gpt-4 technical report.
\newblock 2023.

\bibitem{peng2023instruction}
Baolin Peng, Chunyuan Li, Pengcheng He, Michel Galley, and Jianfeng Gao.
\newblock Instruction tuning with gpt-4.
\newblock {\em arXiv preprint arXiv:2304.03277}, 2023.

\bibitem{radford2021learning}
Alec Radford, Jong~Wook Kim, Chris Hallacy, Aditya Ramesh, Gabriel Goh,
  Sandhini Agarwal, Girish Sastry, Amanda Askell, Pamela Mishkin, Jack Clark,
  et~al.
\newblock Learning transferable visual models from natural language
  supervision.
\newblock In {\em International conference on machine learning}, pages
  8748--8763. PMLR, 2021.

\bibitem{schwenk2022okvqa}
Dustin Schwenk, Apoorv Khandelwal, Christopher Clark, Kenneth Marino, and
  Roozbeh Mottaghi.
\newblock A-okvqa: A benchmark for visual question answering using world
  knowledge.
\newblock In {\em Computer Vision--ECCV 2022: 17th European Conference, Tel
  Aviv, Israel, October 23--27, 2022, Proceedings, Part VIII}, pages 146--162.
  Springer, 2022.

\bibitem{suhr2017corpus}
Alane Suhr, Mike Lewis, James Yeh, and Yoav Artzi.
\newblock A corpus of natural language for visual reasoning.
\newblock In {\em Proceedings of the 55th Annual Meeting of the Association for
  Computational Linguistics (Volume 2: Short Papers)}, pages 217--223, 2017.

\bibitem{touvron2023llama}
Hugo Touvron, Thibaut Lavril, Gautier Izacard, Xavier Martinet, Marie-Anne
  Lachaux, Timoth{\'e}e Lacroix, Baptiste Rozi{\`e}re, Naman Goyal, Eric
  Hambro, Faisal Azhar, et~al.
\newblock Llama: Open and efficient foundation language models.
\newblock {\em arXiv preprint arXiv:2302.13971}, 2023.

\bibitem{xu2022multiinstruct}
Zhiyang Xu, Ying Shen, and Lifu Huang.
\newblock Multiinstruct: Improving multi-modal zero-shot learning via
  instruction tuning.
\newblock {\em arXiv preprint arXiv:2212.10773}, 2022.

\bibitem{zhu2023minigpt}
Deyao Zhu, Jun Chen, Xiaoqian Shen, Xiang Li, and Mohamed Elhoseiny.
\newblock Minigpt-4: Enhancing vision-language understanding with advanced
  large language models.
\newblock {\em arXiv preprint arXiv:2304.10592}, 2023.

\end{thebibliography}

\end{document}